\pdfoutput=1 
\documentclass[10pt,leqno]{amsart}

\usepackage[utf8]{inputenc}   
\usepackage[T1]{fontenc}
\usepackage{lmodern}
\usepackage{microtype}

\usepackage{amsmath,amssymb,amsthm}

\usepackage{graphicx}
\usepackage{xcolor}
\usepackage{booktabs}
\usepackage{multirow}
\usepackage{tabularx}
\usepackage{colortbl}
\usepackage{stfloats}  

\usepackage{algorithm}
\usepackage{algpseudocode}
\usepackage{tikz}
\usetikzlibrary{arrows.meta}
\usepackage{pgfplots}
\pgfplotsset{compat=1.18}

\usepackage{indentfirst}
\usepackage{csquotes}
\usepackage{fancyhdr}

\usepackage[normalem]{ulem}  

\usepackage{hyperref}
\hypersetup{
  colorlinks=true,
  linkcolor=black,
  citecolor=black,
  urlcolor=black
}

\usepackage[margin=1in]{geometry}

\title[KFCPO: Kronecker-Factored Approximated Constrained Policy Optimization]%
{KFCPO: Kronecker-Factored Approximated Constrained Policy Optimization}

\author{Joonyoung Lim}
\address{School of Computer Science and Engineering, Pusan National University, Busan, Korea}
\email{example@mail.com}

\author{Younghwan Yoo}
\address{School of Computer Science and Engineering, Pusan National University, Busan, Korea}
\email{ymomo@pusan.ac.kr}
\thanks{Corresponding author: Younghwan Yoo (\href{mailto:ymomo@pusan.ac.kr}{ymomo@pusan.ac.kr}).}

\date{\today}

\begin{document}
\makeatletter
\let\@setaddresses\relax
\makeatother
\maketitle

\let\thefootnote\relax

\begin{abstract}
We propose KFCPO, a novel Safe Reinforcement Learning (Safe RL) algorithm that combines scalable Kronecker-Factored Approximate Curvature (K-FAC) based second-order policy optimization with safety-aware gradient manipulation. KFCPO leverages K-FAC to perform efficient and stable natural gradient updates by approximating the Fisher Information Matrix (FIM) in a layer-wise, closed-form manner, avoiding iterative approximation overheads. To address the trade-off between reward maximization and constraint satisfaction, we introduce a margin-aware gradient manipulation mechanism that adaptively adjusts the influence of reward and cost gradients based on the agent’s proximity to safety boundaries. This method blends gradients using a direction-sensitive projection, eliminating harmful interference and avoiding abrupt changes caused by fixed hard thresholds. Additionally, a minibatch-level KL rollback strategy is adopted to ensure trust region compliance and to prevent destabilizing policy shifts. Experiments on Safety Gymnasium using OmniSafe show that KFCPO achieves 10.3\%--50.2\% higher average return across environments compared to the best baseline that respected the safety constraint, demonstrating superior balance of safety and performance.
\end{abstract}

\section{Introduction}\label{sec:intro}

Reinforcement Learning (RL) has achieved remarkable success across a variety of domains, including robotics, autonomous systems, and industrial control. However, most RL algorithms remain confined to simulation due to the risks associated with unsafe behavior during training or deployment in real-world environments. This has led to increasing interest in Safe Reinforcement Learning (Safe RL), also known as constrained Reinforcement learning, which seeks to maximize expected return while adhering to predefined safety constraints, typically formulated as cumulative cost thresholds.

To address safety concerns, many Safe RL algorithms have been proposed. Achiam et al.~\cite{CPO} introduced Constrained Policy Optimization (CPO), which applies trust region updates with second order approximations to enforce safety constraints. Subsequent works, including Projected Constrained Policy Optimization (PCPO) by Yang et al.~\cite{yang2020projection}, extended the CPO method through projection-based techniques. Meanwhile, Ray et al.~\cite{ray2019benchmarking} explored Lagrangian-based methods, notably PPO-Lag and TRPO-Lag, which integrate cost constraints into the reward objective via dual variables. These methods build upon PPO~\cite{schulman2017proximal} and TRPO~\cite{schulman2015trust}, two widely used first-order and second order policy optimization algorithms, respectively. While Lagrangian methods are attractive for their computational simplicity and compatibility with existing RL frameworks, CPO and its variants offer stronger theoretical guarantees for constraint satisfaction and more precise control over policy updates. This makes second order approaches particularly appealing in complex, safety-critical scenarios, when computational resources are sufficient.

However, recent benchmark evaluations by Ji et al.~\cite{ji2023safety,ji2024omnisafe} and Ray et al.~\cite{ray2019benchmarking} have shown that CPO and similar second order methods often fail to enforce constraints reliably in high dimensional or partially observable environments. This has revealed two key challenges in constrained RL: (1) approximation errors introduced by the use of conjugate gradient based approximations with limited iterations, and (2) the difficulty of dynamically balancing reward maximization and safety enforcement when the objectives conflict. These findings contradict the results reported in the original CPO paper~\cite{CPO} and emphasize the need for new approaches that better address both optimization accuracy and reward-safety coordination.

Addressing the first challenge optimization accuracy, requires a stable and efficient second order optimizer. To this end, we examine the Kronecker-Factored Approximate Curvature (K-FAC) method, which approximates the Fisher Information Matrix (FIM) using structured Kronecker products to achieve efficient computation with reduced complexity and faster processing.
Originally introduced by Martens and Grosse~\cite{martens2015optimizing}, K-FAC has been extended to support convolutional~\cite{grosse2016kronecker} and recurrent layers~\cite{martens2018kronecker}, enabling its application to a wide range of neural architectures. K-FAC was later extended to reinforcement learning by Wu et al.~\cite{wu2017scalable}, who applied it in the Actor-Critic using Kronecker Factored Trust Region (ACKTR) algorithm. Their work demonstrated that K-FAC not only improves training stability and sample efficiency but also provides strong experimental evidence of its effectiveness when applied to reinforcement learning. However, to the best of our knowledge, K-FAC has not yet been applied in the context of Safe RL.

For the second challenge, reward-safety coordination, researchers have recently investigated gradient manipulation techniques for coordinating conflicting optimization signals. Yu et al.~\cite{yu2020gradient} proposed gradient surgery algorithm as Projecting Conflicting Gradient (PCGrad), which reduces interference by projecting each task’s gradient onto the normal plane of others in multi-task learning. Building on this direction, Gu et al.~\cite{gu2024balance} proposed a soft switching gradient manipulation method for Safe RL, which blends reward and cost gradients by removing conflicting components before combining them. However, their method applies fixed blending ratios within a predefined slack region, making it insensitive to the agent’s actual safety state. Additionally, the introduction of extra hyperparameters (e.g., slack size) increases tuning complexity and reduces robustness across environments.

Building on these insights, we propose KFCPO, a novel Safe RL algorithm that integrates K-FAC for scalable and stable second order optimization with a safety margin aware gradient manipulation mechanism. Our method dynamically adjusts the optimization direction based on the agent’s proximity to constraint violation, avoiding reliance on hard thresholds or manually tuned slack regions. To further improve stability, we introduce a minibatch level Kullback-Leibler (KL) divergence rollback strategy to prevent unexpected policy shifts and preserve trust region behavior. Together, these components enable effective and stable learning in complex, safety-constrained, and partially observable environments.

Our contributions are summarized as follows:
\begin{itemize}
    \item We introduce KFCPO, the first Safe RL algorithm that incorporates K-FAC into Safe RL. 

    \item We propose a novel margin-aware gradient manipulation mechanism that dynamically balances reward and cost gradients based on the agent's proximity to the safety threshold. 

    \item We conduct experiments on the Safety Gymnasium benchmark using the OmniSafe framework, comparing KFCPO with a wide range of baselines including state-of-the-art Safe RL algorithms. Across all environment, KFCPO achieved 10.3\% to 50.2\% higher average return compared to the best performing baseline that respected the cost limit, demonstrating its superior ability to balance safety and performance.

\end{itemize}

\section{Preliminaries}

In this section, we present the background and notations necessary to describe our Safe RL algorithm, including the problem formulation and the core optimization techniques it builds upon. We briefly review the formulation of Constrained Markov Decision Processes (CMDPs), natural gradient based optimization and K-FAC method, which together form the foundation of our approach.

\subsection{Constrained Markov Decision Process}

RL is commonly formulated as a Markov Decision Process (MDP), defined by the tuple \((\mathcal{S}, \mathcal{A}, P, R, \gamma)\), where \(\mathcal{S}\) is the state space, \(\mathcal{A}\) is the action space, \(P(s'|s,a)\) is the transition probability, \(R(s,a)\) is the reward function, and \(\gamma \in [0,1)\) is the discount factor. The objective is to learn a policy \(\pi(a|s)\) that maximizes the expected discounted return:
\begin{equation}
    J(\pi) = \mathbb{E}_{\tau \sim \pi} \left[ \sum_{t=0}^\infty \gamma^t R(s_t, a_t) \right].
\end{equation}

To incorporate safety constraints, this framework is extended to a CMDPs, introduced by Altman~\cite{altman2021constrained}, where a cost function \(C(s,a)\) quantifies constraint violations and a threshold \(d\) bounds the expected cumulative cost. The constrained objective becomes:
\begin{equation}
    \max_\pi J(\pi) \quad \text{s.t.} \quad J_C(\pi) = \mathbb{E}_{\tau \sim \pi} \left[ \sum_{t=0}^\infty \gamma^t C(s_t, a_t) \right] \leq d.
\end{equation}

This constrained formulation introduces a trade-off between maximizing performance and satisfying safety requirements.

\subsection{Natural Gradient}

Sutton et al.~\cite{sutton1999policy} proposed the use of policy gradients to optimize the parameters \(\theta\) of a policy \(\pi_\theta(a|s)\) by following the direction of the first-order gradient \(\nabla_\theta J(\pi)\). While effective in many settings, this approach ignores second order information, notably curvature, potentially leading to inefficient or unstable learning.

To address this issue, Amari~\cite{amari1998natural} introduced the concept of the natural gradient, which incorporates second order information by preconditioning the gradient with the inverse of the FIM. The FIM, as formalized by the work of Kakade~\cite{kakade2001natural}, is defined as:
\begin{equation}
    F(\theta) = \mathbb{E}_{s \sim d^\pi, a \sim \pi_\theta} \left[ \nabla_\theta \log \pi_\theta(a|s) \nabla_\theta \log \pi_\theta(a|s)^\mathrm{T} \right].
\end{equation}
The natural gradient is given by:
\begin{equation}
    \tilde{\nabla}_\theta J(\pi) = F(\theta)^{-1} \nabla_\theta J(\pi).
\end{equation}
For notational convenience, we denote the standard policy gradient as \( g = \nabla_\theta J(\pi) \) and its natural gradient as \( \tilde{g} = \tilde{\nabla}_\theta J(\pi) \). This shorthand will be used throughout the rest of the paper.

Schulman et al.~\cite{schulman2015trust} applied this concept in trust region methods including TRPO, where the natural gradient led to more stable and efficient policy updates. However, computing and inverting the FIM can be computationally expensive, particularly in large policy networks.

\subsection{Kronecker-Factored Approximate Curvature}

As discussed in Section~\ref{sec:intro}, K-FAC is a scalable second order optimization method that approximates the FIM using Kronecker products. This structure exploits the layered nature of neural networks, enabling efficient natural gradient computation even in high dimensional models~\cite{martens2015optimizing, grosse2016kronecker, martens2018kronecker}.

For a fully connected layer with weight matrix \(W\), the corresponding FIM block is approximated as:
\begin{equation}
    F_W \approx A \otimes G,
\end{equation}
where \(A = \mathbb{E}[aa^\mathrm{T}]\) represents the covariance of layer inputs and \(G = \mathbb{E}[\nabla_W \log \pi \nabla_W \log \pi^\mathrm{T}]\) denotes the covariance of output gradients. The inverse is approximated analytically as:
\begin{equation}
    F_W^{-1} \approx A^{-1} \otimes G^{-1}.
\end{equation}

This Kronecker factorization reduces the computational cost of inverting the FIM by approximating the full parameter matrix with layer wise smaller matrices, improving scalability for large neural networks. It also enables parallelized computation across layers and facilitates memory efficient storage, making K-FAC suitable for deep networks with complex architectures.

In RL, Wu et al.~\cite{wu2017scalable} employed K-FAC within the ACKTR algorithm, demonstrating that it yields more stable policy updates and improved sample efficiency compared to conjugate gradient methods. Unlike iterative solvers, K-FAC provides a structured and closed-form approximation of the natural gradient, enabling practical second order updates even in high dimensional policy networks.

\section{Proposed Method}

In this section, we introduce \textbf{KFCPO}, a Safe RL algorithm that integrates K-FAC based natural gradient estimation with constrained policy optimization. KFCPO is designed to ensure stable and efficient learning under safety constraints. The following subsections describe its key components, followed by an overview of the full algorithm.

\subsection{Utilization of K-FAC for Policy Optimization}

To enable scalable and stable policy optimization under safety constraints, we incorporate K-FAC into our algorithm. Specifically, we maintain separate FIM approximations for the reward and cost objectives, enabling dual objective natural gradient updates.

For each linear layer in the policy network, we estimate the input activation covariance \(A\) and the output gradient covariance \(G\) using exponential moving averages:
\begin{equation}
A(s) = (1 - \epsilon) A(s-1) + \epsilon A_{\text{new}}, \quad 
G(s) = (1 - \epsilon) G(s-1) + \epsilon G_{\text{new}},
\end{equation}
where the decay factor \(\epsilon = 0.95\), and \(A_{\text{new}}, G_{\text{new}}\) are computed from the current minibatch. Here, \(s\) denotes the minibatch update step at which K-FAC statistics are incrementally updated.

To efficiently invert the approximated FIM, we perform eigendecomposition:
\begin{equation}
A = Q_A \Lambda_A Q_A^\mathrm{T}, \quad G = Q_G \Lambda_G Q_G^\mathrm{T},
\end{equation}
\begin{equation}
F^{-1} \approx (Q_A \Lambda_A^{-1} Q_A^\mathrm{T}) \otimes (Q_G \Lambda_G^{-1} Q_G^\mathrm{T}).
\end{equation}
Here, \(Q_A\) and \(Q_G\) are orthogonal matrices whose columns are the eigenvectors of \(A\) and \(G\), respectively, and \(\Lambda_A\) and \(\Lambda_G\) are diagonal matrices containing the corresponding eigenvalues. This eigendecomposition enables efficient and numerically stable inversion of each Kronecker factor.

To ensure that updates remain within a trust region, we compute a KL-based scaling factor adjusted for minibatch size:
\begin{equation}
    \nu = \min\left( \nu_{\text{max}}, \, \frac{|\mathcal{B}|}{N} \cdot \sqrt{\frac{2 \delta}{\tilde{g}^\mathrm{T} F \tilde{g}}} \right),
    \label{eq:kl_scaling}
\end{equation}
where \(\delta\) is the target KL divergence bound, and \(\tilde{g}\) denotes the final natural gradient direction after the gradient manipulation process described in Section~\ref{subsec:gradient_manipulation}. The coefficient \(\frac{|\mathcal{B}|}{N}\) represents the ratio between minibatch size and the total number of samples in the epoch. This adjustment ensures that the trust region scaling remains valid when using minibatches, by proportionally reducing the update magnitude based on the minibatch to epoch size ratio.

The final update is performed using momentum:
\begin{equation}
m(t) = \beta m(t-1) + \nu \cdot \tilde{g}(t), \quad 
\theta(t+1) = \theta(t) - \alpha \cdot m(t)
\end{equation}
The effective learning rate \(\alpha = lr \cdot (1 - \beta)\), where \(\beta\) is a momentum coefficient.
\subsection{Safety Margin-Based Gradient Manipulation} \label{subsec:gradient_manipulation}

To maintain long term safety without sacrificing reward, we adaptively blend the reward and cost natural gradients based on the agent’s current safety state. When the policy is unsafe, we prioritize cost. When it is near the constraint, we bias the update toward cost. In clearly safe cases, reward optimization dominates with minimal constraint interference.

Let \(C\) denote the cost limit and \(c_{\text{ep}}\) the average episodic cost. We define the center of the safety margin as \(C_{\text{center}} = \lambda C\), where \(\lambda \in (0,1)\) is a predefined margin coefficient. The blending weights are computed as:
\begin{equation}
    w_c = \frac{1}{1 + \exp\left(-k(c_{\text{ep}} - C_{\text{center}})\right)}, \quad w_r = 1 - w_c,
\end{equation}
where \(k\) controls the steepness of the transition.

\begin{figure}[t]
    \centering
    \includegraphics[width=0.9\linewidth]{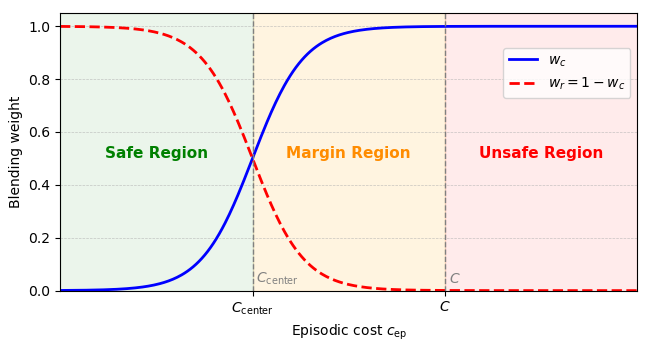}
    \caption{Adaptive blending weights based on $c_{\text{ep}}$.}
    \label{fig:blending_ratio}
\end{figure}

Figure~\ref{fig:blending_ratio} illustrates how the blending weights \(w_r\) and \(w_c\) evolve as a function of the average episodic cost \(c_{\text{ep}}\). The X-axis is divided into three regions: \textit{Safe}, \textit{Margin}, and \textit{Unsafe}, separated by \(C_{\text{center}}\) and \(C\). In the \textit{safe zone} (\(c_{\text{ep}} < C_{\text{center}}\)), \(w_r\) remains high while \(w_c\) is close to zero, allowing reward gradients to dominate. As \(c_{\text{ep}}\) approaches \(C_{\text{center}}\), the agent enters the \textit{margin zone}, and \(w_c\) starts to increase sharply, reflecting growing concern for constraint satisfaction. Beyond the constraint threshold \(C\), in the \textit{unsafe zone}, \(w_c\) saturates near 1 while \(w_r\) becomes negligible, effectively prioritizing the cost gradient.

To handle potential conflict between the reward and cost gradients, we adopt a direction aware blending strategy inspired by the PCGrad proposed by Yu et al.~\cite{yu2020gradient} in the context of multi-task learning. When the angle between \(\tilde{g}_r\) and \(\tilde{g}_c\) is less than 90 degrees (i.e., $\cos(\tilde{g}_r, \tilde{g}_c) > 0$), the two gradients are directly blended:
\begin{equation}
    \tilde{g}(t) = w_r \tilde{g}_r + w_c \tilde{g}_c.
\end{equation}

When the gradients point in opposing directions (i.e., $\cos(\tilde{g}_r, \tilde{g}_c) \leq 0$), we remove the component of the cost gradient that interferes with the reward direction using orthogonal projection:
\begin{equation}\label{eq:g_c_top}
    \tilde{g}_c^{\perp} = \tilde{g}_c - \frac{\tilde{g}_c^\mathrm{T} \tilde{g}_r}{\|\tilde{g}_r\|^2 + \epsilon} \tilde{g}_r,
\end{equation}
and the final update becomes:
\begin{equation}
    \tilde{g}(t) = w_r \tilde{g}_r + w_c \tilde{g}_c^{\perp}.
\end{equation}

\begin{figure}[t]
    \centering
    \includegraphics[width=0.9\linewidth]{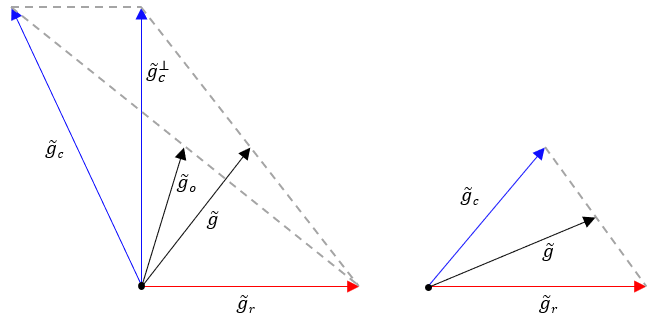}
    \caption{Blending behavior: (left) with projection, (right) direct combination.}
    \label{fig:surgery}
\end{figure}

Figure~\ref{fig:surgery} illustrates the geometric behavior of our blending strategy. In the left part of Figure~\ref{fig:surgery}, the angle between $\tilde{g}_r$ and $\tilde{g}_c$ exceeds 90 degrees, indicating a directional conflict. In this case, we compute the projection of $\tilde{g}_c$ by removing its component in the direction of $\tilde{g}_r$, as described in the Equation (\ref{eq:g_c_top}) for $\tilde{g}_c^\perp$. The final update $\tilde{g}$ is then obtained by blending $\tilde{g}_r$ and $\tilde{g}_c^\perp$ with adaptive weights. The vector $\tilde{g}_o$ shown in the figure represents the naively blended direction without projection, allowing a direct comparison against our method’s projection-based result. In the right part of the Figure~\ref{fig:surgery}, the angle between the gradients is less than 90 degrees, and thus they can be blended directly without projection.

\begin{figure}[t]
    \centering
    \includegraphics[width=0.9\linewidth]{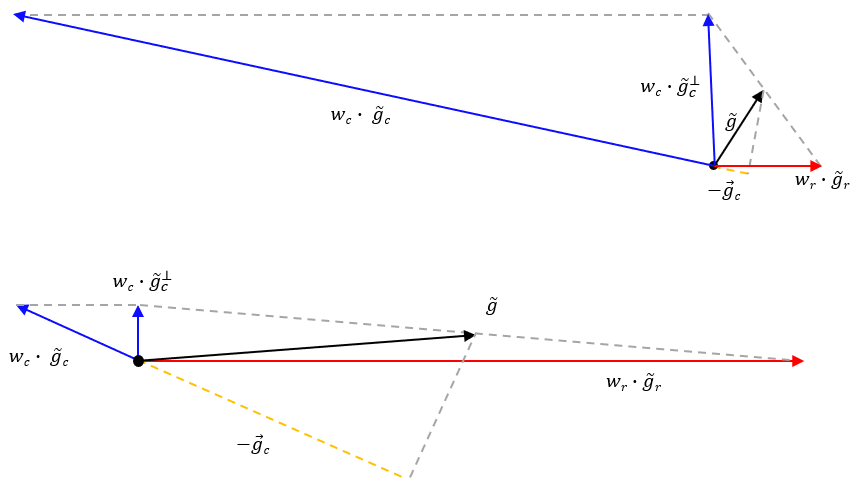}
    \caption{Final updates under conflict show varying degrees of misalignment with $\tilde{g}_c$.}
    \label{fig:surgery_vs_margin}
\end{figure}

While projection resolves direct conflicts, Gu et al.~\cite{gu2024balance} point out a limitation in the concept of PCGrad. Specifically, when the angle between $\tilde{g}_r$ and $\tilde{g}_c$ exceeds 90 degrees, the final update direction $\tilde{g}$ can retain a component that is negatively aligned with the original cost gradient $\tilde{g}_c$, denoted as $-\tilde{g}_c$. This implies that projection alone cannot fully eliminate conflict when $\tilde{g}_r$ and $\tilde{g}_c$ point in nearly opposite directions, which can lead to constraint degradation or instability.

However, our proposed blending mechanism mitigates this issue by adaptively scaling the influence of each objective based on the policy’s safety context. Figure~\ref{fig:surgery_vs_margin} illustrates how the final update direction behaves under such high-conflict scenarios, after applying the blending weights $w_r$ and $w_c$. In the upper part of Figure~\ref{fig:surgery_vs_margin}, the policy lies within the margin zone, where the cost weight $w_c$ is significantly larger than $w_r$. Although the final update direction retains a component opposite to the cost gradient (i.e., $-\tilde{g}_c$, shown as a yellow dashed vector), this component remains small due to the projection and adaptive weighting. This allows the update to preserve performance while still accounting for the influence of the cost gradient. In the lower part of the Figure~\ref{fig:surgery_vs_margin}, the policy is in the safe zone, and the reward weight $w_r$ dominates. In this case, the update direction includes a non-negligible component in the direction of $-\tilde{g}_c$, but since the policy remains well within the safety constraint, this trade-off is acceptable for reward maximization without risking constraint violation.

Importantly, in both the safe and margin regions, the average cost remains below the safety threshold. The update direction naturally adjusts to the agent's safety status, favoring reward optimization in safe regions and increasing sensitivity to constraints near the boundary. This leads to stable, efficient, and safety preserving policy improvement without relying on short term performance gains.

\subsection{minibatch level KL Rollback}

To ensure stable policy improvement, we incorporate a minibatch level KL rollback mechanism inspired by early stopping strategies in trust region methods. As introduced in Equation~(\ref{eq:kl_scaling}), the natural gradient update is scaled to satisfy a target KL bound. However, some minibatches may still produce overly aggressive updates. To prevent this, we compute the KL divergence between the updated policy $\pi_{\text{new}}$ and the previous policy $\pi_{\text{old}}$ after each minibatch update. If the divergence exceeds a predefined threshold $\delta$, the update is rolled back to ensure trust region compliance.

Moreover, the original TRPO and PPO papers, as well as many implementation works that build on them, typically adopt KL divergence limits in the range of 0.01 to 0.02~\cite{schulman2015trust, schulman2017proximal,stable-baselines3}.
 we deliberately use a more conservative threshold of $\delta$ = 0.005 to 0.01 in KFCPO. This stricter constraint reduces the likelihood of unsafe or unstable policy shifts, particularly in high dimensional and noisy gradient settings. Although it may lead to slower convergence compared to more aggressive approaches, the resulting updates are significantly more stable and safety-preserving. This trade-off aligns with the KFCPO’s core concept: prioritizing reliable constraint satisfaction over rapid reward improvement.

\begin{algorithm}[t]
\caption{KFCPO: Safe Policy Optimization with K-FAC and Safety Aware Gradient Blending}
\label{alg:kfcpo}
\begin{algorithmic}[1]
\Require Initial policy $\pi_\theta$, critics $V_r$, $V_c$
\State \parbox[t]{\dimexpr\linewidth-\algorithmicindent}{
\textbf{Hyperparameters:} learning rate $\alpha$, damping $\delta$, cost limit $C$, margin coefficient $\lambda$, trust region bound $\epsilon_{\text{KL}}$, sigmoid scale $k$, K-FAC intervals $T_s$, $T_f$, batch size $B$, update steps $K$
}
\State \parbox[t]{\dimexpr\linewidth-\algorithmicindent}{
Initialize Layerwise K-FAC optimizer with momentum
}
\For{each epoch}
    \State \parbox[t]{\dimexpr\linewidth-\algorithmicindent}{
    Collect trajectories using current policy $\pi_\theta$
    }
    \State \parbox[t]{\dimexpr\linewidth-\algorithmicindent}{
    Update critics $V_r$, $V_c$ using temporal-difference targets
    }
    \State \parbox[t]{\dimexpr\linewidth-\algorithmicindent}{
    Compute returns, advantages $A_r$, $A_c$, and log-probs $\log \pi_{\theta_{\text{old}}}$
    }
    \State \parbox[t]{\dimexpr\linewidth-\algorithmicindent}{
    Estimate average episodic cost $c_{\text{ep}}$
    }
    \State \parbox[t]{\dimexpr\linewidth-\algorithmicindent}{
    Compute blending weights: $w_c \gets \sigma\left((c_{\text{ep}} - \lambda C) \cdot k\right)$,\quad $w_r \gets 1 - w_c$
    }
    \For{$k = 1$ to $K$}
        \For{each minibatch $(s, a, \log \pi_{\theta_{\text{old}}}, A_r, A_c)$}
            \State \parbox[t]{\dimexpr\linewidth-\algorithmicindent}{
            Compute standard gradients: $g_r \gets \nabla_\theta J(\pi)$,\\
            \hspace*{11.5em} $g_c \gets \nabla_\theta J_C(\pi)$
            }

            \State \parbox[t]{\dimexpr\linewidth-\algorithmicindent}{
            Update Fisher statistics if $k \bmod T_s = 0$
            }
            \State \parbox[t]{\dimexpr\linewidth-\algorithmicindent}{
            Compute natural gradients: $\tilde{g}_r \gets F_r^{-1} g_r$,\\
            \hspace*{10em} \quad $\tilde{g}_c \gets F_c^{-1} g_c$
            }
            \If{$\cos(\tilde{g}_r, \tilde{g}_c) > 0$}
                \State \parbox[t]{\dimexpr\linewidth-\algorithmicindent}{
                $\tilde{g} \gets w_r \tilde{g}_r + w_c \tilde{g}_c$
                }
            \Else
                \State \parbox[t]{\dimexpr\linewidth-\algorithmicindent}{
                Compute projection: $\tilde{g}_c^\perp \gets \tilde{g}_c - \frac{\tilde{g}_c^\mathrm{T} \tilde{g}_r}{\|\tilde{g}_r\|^2 + \epsilon} \tilde{g}_r$
                }
                \State \parbox[t]{\dimexpr\linewidth-\algorithmicindent}{
                Blended direction: $\tilde{g} \gets w_r \tilde{g}_r + w_c \tilde{g}_c^\perp$
                }
            \EndIf
            \State \parbox[t]{\dimexpr\linewidth-\algorithmicindent}{
            
            Compute scaling factor: \\ \hspace*{7em}
            $\nu \gets \min\left(\nu_{\text{max}}, \frac{|\mathcal{B}|}{N} \cdot \sqrt{\frac{2 \delta}{\tilde{g}^\mathrm{T} F \tilde{g}}} \right)$
            }
            \State \parbox[t]{\dimexpr\linewidth-\algorithmicindent}{
            Construct candidate update: $\theta' \gets \theta - \nu \tilde{g}$
            }
            \State \parbox[t]{\dimexpr\linewidth-\algorithmicindent}{
            Evaluate KL divergence: $\mathrm{KL}(\pi_{\theta'} \| \pi_\theta)$
            }
            \If{$\mathrm{KL}(\pi_{\theta'} \| \pi_\theta) > \epsilon_{\text{KL}}$}
                \State \parbox[t]{\dimexpr\linewidth-\algorithmicindent}{
Roll back: $\theta \gets \theta_{\text{old}}$
}
                \State \textbf{continue}
            \Else
                \State \parbox[t]{\dimexpr\linewidth-\algorithmicindent}{
                Commit update: $\theta \gets \theta'$
                }
            \EndIf
            \If{$k \bmod T_f = 0$}
                \State \parbox[t]{\dimexpr\linewidth-\algorithmicindent}{
                Refresh Fisher eigendecomposition
                }
            \EndIf
        \EndFor
    \EndFor
\EndFor
\end{algorithmic}
\end{algorithm}

\subsection{Algorithm Overview}

KFCPO is an on-policy actor-critic algorithm that integrates second order natural gradient optimization with adaptive gradient blending under safety constraints. The policy is updated using on-policy trajectories, while two separate critics estimate the expected returns and costs to provide advantage estimates for learning.

Algorithm~\ref{alg:kfcpo} outlines the full procedure of KFCPO. At each epoch, the agent collects trajectories using the current policy $\pi_\theta$, and computes the reward and cost advantages $A_r$ and $A_c$ from value estimators $V_r$ and $V_c$, respectively.

The average episodic cost $c_{\text{ep}}$ is used to compute adaptive blending weights $(w_r, w_c)$ through a sigmoid function centered at the safety margin. For each minibatch, policy gradients $g_r = \nabla_\theta J(\pi)$ and $g_c = \nabla_\theta J_C(\pi)$ are computed and preconditioned via the K-FAC method to obtain natural gradients $\tilde{g}_r$ and $\tilde{g}_c$.

A margin aware manipulation mechanism constructs the final update direction $\tilde{g}$, using projection when the gradients are in conflict. The update is scaled by a trust region coefficient $\nu$ based on the KL divergence constraint. Before the policy is updated, the KL divergence between $\pi_\theta$ and the candidate policy $\pi_{\theta'}$ is computed. If the KL divergence exceeds the threshold $\epsilon_{\text{KL}}$, the update is rolled back.

This minibatch level KL rollback ensures stability while allowing for adaptive and efficient policy improvement under safety constraints.

\section{Empirical Evaluation}

\subsection{Experimental Setup}\label{sec:experimental_setup}

We evaluated KFCPO on Safety Gymnasium~\cite{ji2023safety}, a recently proposed and actively maintained benchmark for Safe RL built on the MuJoCo physics engine~\cite{todorov2012mujoco}. Safety Gymnasium unifies and extends previous work, notably OpenAI's Safety Gym~\cite{ray2019benchmarking}, offering standardized environments based on the CMDP formulation. Our experiments were conducted on four single agent environments, each defined by a unique combination of agent type and task type. We use two agent types: \textbf{Point}, a 2D planar mobile agent, and \textbf{Car}, a 3D non-holonomic mobile agent. These are evaluated in two task types: \textbf{Goal}, where the agent must reach a fixed target location, and \textbf{Button}, where the agent must press a highlighted goal button among distractors.

\begin{figure}[t]
    \centering
    \includegraphics[width=0.7\linewidth]{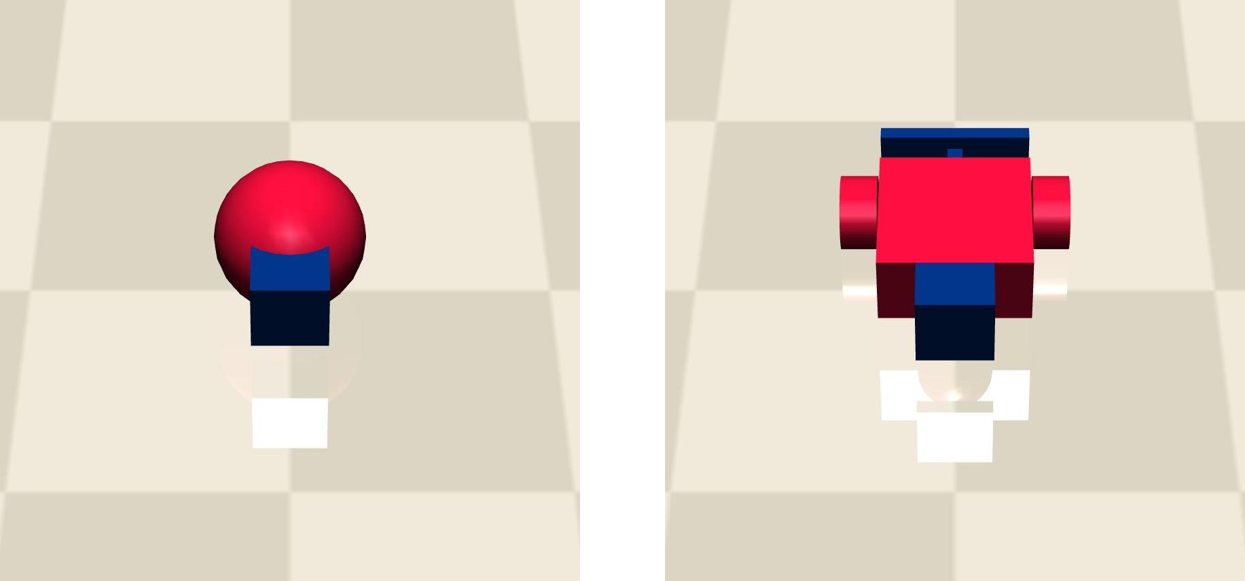}
    \caption{Visual appearances of agents: (left) Point, (right) Car agent.}
    \label{fig:agent_visuals}
\end{figure}

The point agent is a planar robot controlled via forward/backward force and rotational velocity, with a 12 dimensional observation space consisting of motion-related sensors such as an accelerometer, velocimeter, gyroscope, and magnetometer. The car agent uses differential drive by applying independent forces to its left and right wheels and extends the point’s observation space to 24 dimensions by incorporating rear wheel orientation and angular velocity. Figure~\ref{fig:agent_visuals} show the visual appearance of the point and car agents.

\begin{figure}[t]
    \centering
    \includegraphics[width=0.7\linewidth]{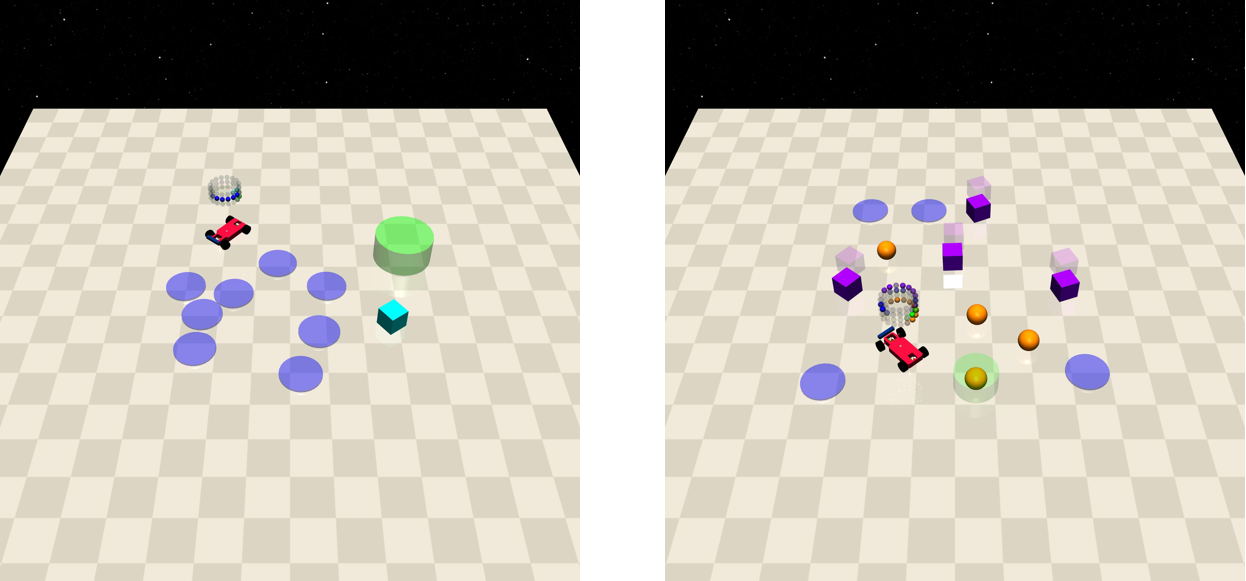}
    \caption{Examples of task environments: (left) Goal, (right) Button task.}
    \label{fig:task_visuals}
\end{figure}

In both tasks, the agent must interact with designated targets while avoiding unsafe interactions. In goal tasks, the agent receives sparse rewards upon entering the goal region and dense rewards based on distance to the goal. Hazards act as active constraints, and vases serve as non-penalizing distractors. As shown in left part of Figure~\ref{fig:task_visuals}\,, the agent's objective is to reach the green colored circular goal region while avoiding hazards. In button tasks, the agent must press a highlighted goal button among several orange distractors. Costs are incurred for contacting hazards, colliding with gremlins, or pressing the wrong button. The right side of Figure~\ref{fig:task_visuals}\, illustrates the layout, where only the green-highlighted button yields a reward.

Table~\ref{tab:agent_task_detailed} summarizes the agent-task combinations, including action and observation structures, lidar based inputs, and associated constraints.

\begin{table*}[t]
\centering
\caption{Agent-task configurations in Safety-Gymnasium: Action space, observation structure, task input, and constraints.}
\label{tab:agent_task_detailed}
\resizebox{\textwidth}{!}{
\begin{tabular}{|l|l|l|l|l|l|}
\hline
\textbf{Agent} & \textbf{Action Space} & \textbf{Agent Observation Space} & \textbf{Task} & \textbf{Task Observation Space} & \textbf{Constraints} \\
\hline
\multirow{1}{*}{\textbf{Point}}  
& \begin{tabular}[t]{@{}l@{}}2D: forward/backward force, rotation\end{tabular} 
& \begin{tabular}[t]{@{}l@{}}12D: accelerometer, velocimeter, \\ gyroscope, magnetometer\end{tabular}
& \textbf{Goal} 
& \begin{tabular}[t]{@{}l@{}}48D: goal, hazard, vase lidar (16×3)\end{tabular}
& \begin{tabular}[t]{@{}l@{}}Avoid 8 hazards \\ Vases (no cost)\end{tabular} \\
\hline

\multirow{1}{*}{\textbf{Car}}  
& \begin{tabular}[t]{@{}l@{}}2D: left/right wheel force\end{tabular} 
& \begin{tabular}[t]{@{}l@{}}24D: rear wheel sensors, \\ accelerometer, velocimeter, \\ gyroscope, magnetometer\end{tabular}
& \textbf{Button} 
& \begin{tabular}[t]{@{}l@{}}64D: button, goal, gremlin, hazard lidar (16×4)\end{tabular}
& \begin{tabular}[t]{@{}l@{}}Avoid 4 hazards, 4 gremlins \\ Penalized for wrong button\end{tabular} \\
\hline
\end{tabular}
}
\end{table*}

All agents were trained for 1 million environment steps using an on-policy actor-critic framework with separate value networks for reward and cost. We used generalized advantage estimation (GAE)~\cite{schulman2015high} with $\gamma = 0.99$ and $\lambda = 0.97$ to compute the advantage estimates. K-FAC updates are performed every $T_s = 1$ step, with eigendecomposition refreshed every $T_f = 10$ steps. A trust region is enforced using a KL-divergence threshold of $\delta$ = 0.005, and the KL rollback mechanism is applied at the minibatch level to ensure stable updates.

\textbf{To ensure a fair comparison across all algorithms}, we adopted the default hyperparameter settings provided by the OmniSafe framework~\cite{ji2024omnisafe}. All baseline methods are implemented and executed using their official OmniSafe configurations. Moreover, all policy and value networks share the same architecture across experiments: a two-layer multilayer perceptron (MLP) with 64 hidden units per layer and ReLU activations. This standardization isolates the performance differences to the algorithmic level, eliminating potential confounding factors due to architecture or tuning. All experiments were repeated across three random seeds. We report the average episodic return and average constraint cost over the final 10 epoch.

\subsection{Benchmark Algorithms}
To evaluate performance in these environments, we compare KFCPO against Safe RL baselines: CPO~\cite{CPO} and PCPO~\cite{yang2020projection} as second order constrained optimization methods; TRPO-Lag and PPO-Lag~\cite{ray2019benchmarking} as Lagrangian dual approaches; and two SOTA algorithms, \textit{Constrained Update Projection} (CUP)~\cite{yang2022constrained} and \textit{Penalized Proximal Policy Optimization} (P3O)~\cite{zhang2022penalized}.

CUP is a method proposed by Yang et al.~\cite{yang2022constrained}, which leverages GAE-based non-convex optimization to decouple policy improvement from constraint enforcement. It provides theoretically guaranteed safe updates within a trust region optimization framework and is designed for stable and scalable learning under complex CMDP conditions.

P3O is an algorithm introduced by Zhang et al.~\cite{zhang2022penalized}, which reformulates the constrained optimization problem as an unconstrained one using ReLU-based penalty functions. It incorporates clipped surrogate objectives, similar to PPO, to achieve sample-efficient and safe learning in high dimensional tasks.

\begin{figure*}[t]
    \centering
    \includegraphics[width=0.9\textwidth]{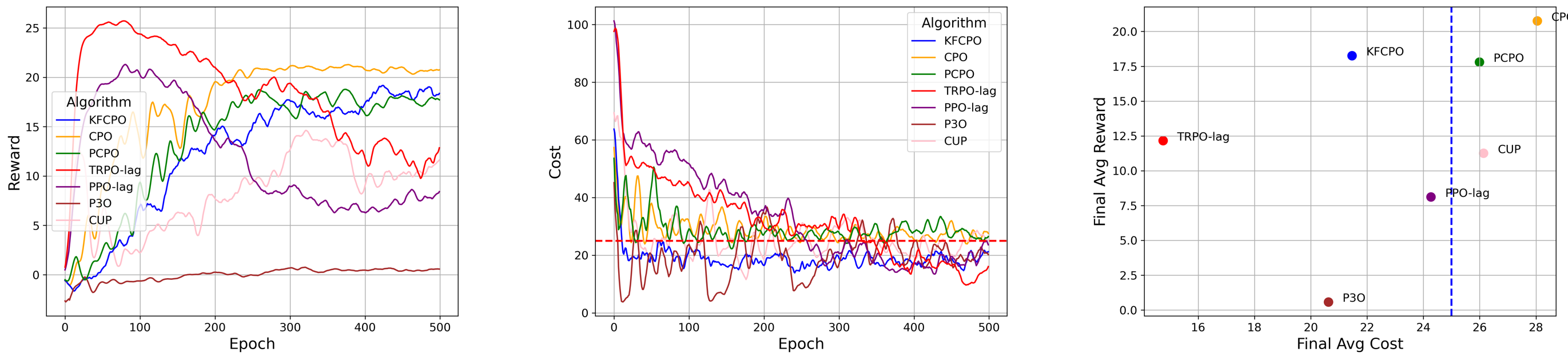}
    \caption{Evaluation on SafetyPointGoal: Reward and cost over training epochs, followed by the final reward–cost outcome.}
    \label{fig:SafetyPointGoal}

    \includegraphics[width=0.9\textwidth]{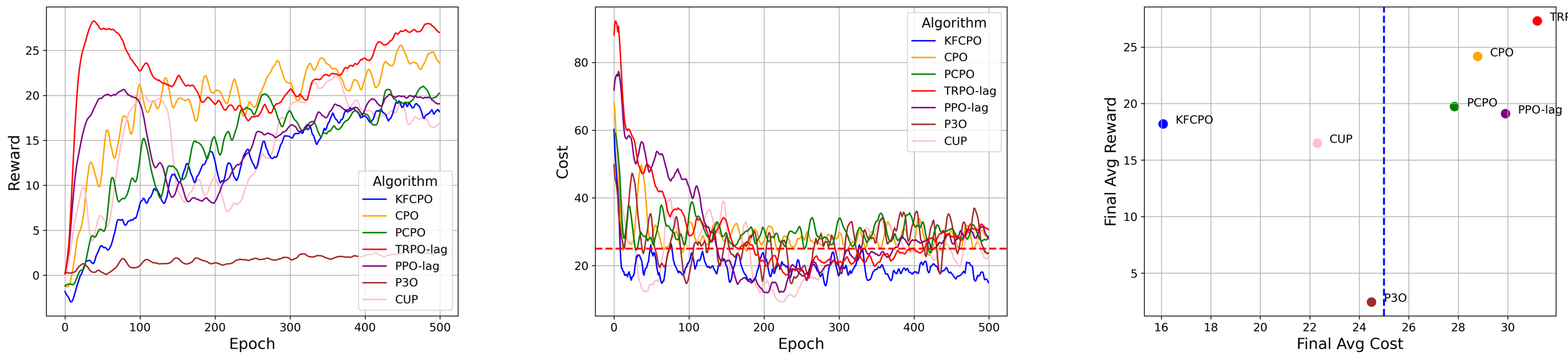}
    \caption{Evaluation on SafetyCarGoal: Reward and cost over training epochs, followed by the final reward–cost outcome}
    \label{fig:SafetyCarGoal}
\end{figure*}

\subsection{Experimental Results}

\begin{figure*}[t]
    \centering
    \includegraphics[width=0.9\textwidth]{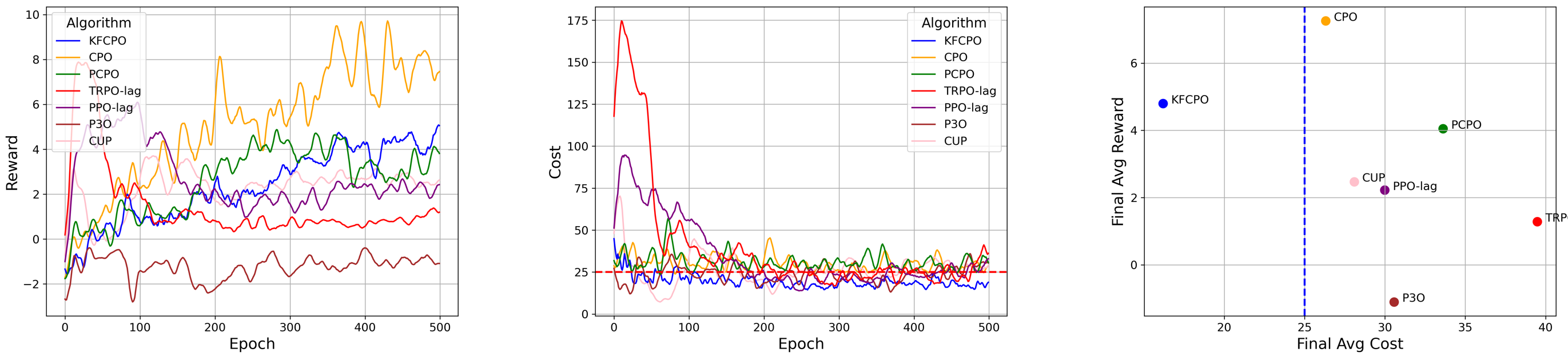}
    \caption{Evaluation on SafetyPointButton: Reward and cost over training epochs, followed by the final reward–cost outcome.}
    \label{fig:SafetyPointButton}

    \includegraphics[width=0.9\textwidth]{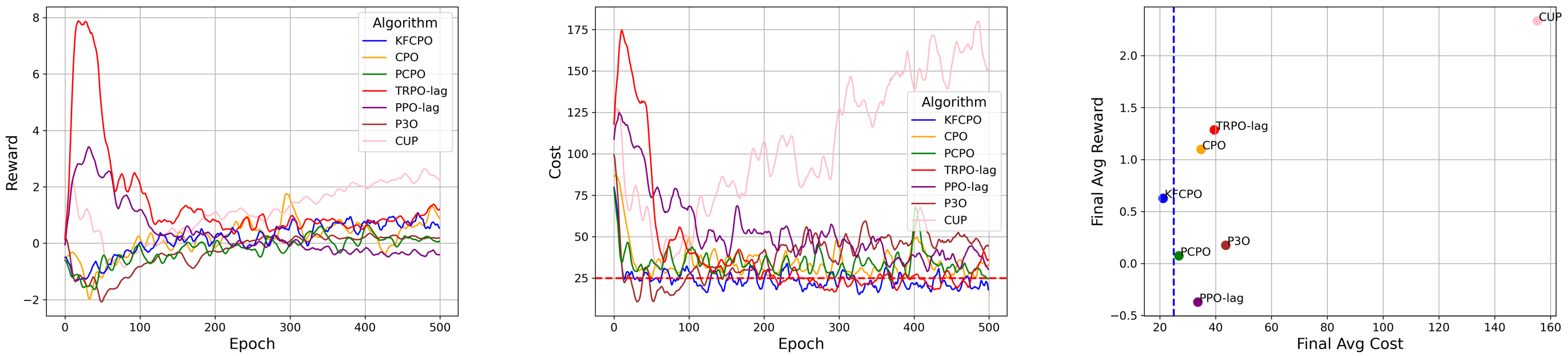}
    \caption{Evaluation on SafetyCarButton: Reward and cost over training epochs, followed by the final reward–cost outcome.}
    \label{fig:SafetyCarButton}
\end{figure*}

As described in Section~\ref{sec:experimental_setup}, we evaluate our method on four distinct combinations of agent and task types: \textbf{SafetyPointGoal}, \textbf{SafetyCarGoal}, \textbf{SafetyPointButton}, and \textbf{SafetyCarButton}. These names correspond to combinations of the Point or Car agent with either the Goal or Button task. The evaluation results are presented in Figures~\ref{fig:SafetyPointGoal}, \ref{fig:SafetyCarGoal}, \ref{fig:SafetyPointButton}, and \ref{fig:SafetyCarButton}, respectively. Each figure presents three horizontally aligned plots. From left to right, they show: (1) the epoch-based reward learning curve, (2) the epoch-based cost curve, and (3) the final trade-off between reward and cost.

In the SafetyPointGoal environment, among the algorithms that respected the cost limit, KFCPO achieved the highest reward, outperforming TRPO-Lag by 50.2\% and PPO-Lag by 125\%. P3O also stayed within the cost limit but is excluded from meaningful performance comparison due to unstable learning behavior in this and subsequent environments. CPO recorded the highest overall reward but exceeded the cost limit with an average cost of 28.05. In contrast, KFCPO maintained a cost of 21.46, staying safely within the defined cost limit.

In the SafetyCarGoal environment, KFCPO, CUP, and P3O satisfied the cost constraint, with KFCPO achieving the highest reward among them. Specifically, KFCPO outperformed CUP by 10.3\% in average return, while maintaining a lower average cost (16.07 vs. 22.29). All other baselines, including CPO, PCPO, and TRPO-Lag, failed to remain within the cost limit. KFCPO maintained a stable cost throughout training, demonstrating strong constraint adherence while preserving competitive return.

These results are consistent with prior studies~\cite{ray2019benchmarking, ji2023safety, ji2024omnisafe} and support the limitations discussed in Section~\ref{sec:intro}. One reason of the constraint violations in CPO and PCPO is the approximation error introduced by iterative conjugate gradient procedures. KFCPO overcomes this limitation by employing K-FAC, which computes second order updates in a stable and analytical form, leading to improved adherence to safety constraints.

The learning curves for reward and cost further highlight behavioral differences across algorithms. Lagrangian-based methods, TRPO-Lag and PPO-Lag, exhibit fast reward improvement in the early stages of training, but this comes at the expense of prolonged cost violations due to slow adaptation of the penalty signal. In contrast, KFCPO adopts a more conservative strategy in the initial training phase, prioritizing the cost gradient to address unsafe behaviors early on. This allows the average cost to quickly fall below the safety threshold. Once within the safe or margin zones, a zone-aware blending mechanism adaptively balances reward and cost gradients based on safety status. As a result, KFCPO consistently maintains safety after the first 10 epochs with low variance in both reward and cost. While this cautious approach may slow convergence compared to more aggressive methods, it ensures stable and reliable learning, which is an essential requirement for safe real-world deployment. This behavior reflects the core goal of Safe RL, which is to enable agents to avoid unsafe actions not only during training but also during deployment in real-world environments.

In the SafetyPointButton and SafetyCarButton environments, as illustrated in Figure~\ref{fig:SafetyPointButton} and \ref{fig:SafetyCarButton}, KFCPO was the only algorithm that consistently satisfied the cost constraint, demonstrating strong robustness even in complex and noisy safety-critical scenarios. Other baselines failed to meet the safety requirement in both environments, highlighting the difficulty of constraint enforcement under increased task complexity and observation dimensionality.

The relatively low reward performance across all methods in the Button tasks can be attributed to the increased complexity of the observation space. As summarized in Table~\ref{tab:agent_task_detailed}, for the Point agent, the observation dimension increases from 60 in the Goal tasks to 76 in the Button tasks. Similarly, for the Car agent, the dimension increases from 72 to 88. Despite this substantial growth in input dimensionality, all algorithms were trained under the default network configuration described in Section~\ref{sec:experimental_setup}, where a two-layer MLP with 64 hidden units per layer was used. Consequently, the input size exceeded the size of each hidden layer, limiting the network's expressiveness and its ability to accurately model the complex dynamics required for successful task completion. This architectural constraint likely contributed to the reduced reward performance observed in the Button tasks.

Despite these architectural limitations, KFCPO consistently satisfied the cost constraint across both Button environments, whereas all other baselines failed to do so. This robustness under suboptimal model capacity highlights a significant advantage for real-world deployment. In practical settings, an inappropriate model structure or insufficient network capacity can easily lead to severe constraint violations, which in turn may cause catastrophic failures or significant financial loss. However, by effectively suppressing excessive cost accumulation even under constrained model conditions, KFCPO enables systems to maintain operational safety and facilitates further model refinement through continued learning. Consequently, rather than requiring system shutdowns or model replacement, KFCPO supports safe and incremental improvement, offering a considerable practical advantage for deploying reinforcement learning agents in complex and safety-critical settings.

\section{Conclusion}

This work introduced KFCPO, a novel Safe RL algorithm that integrates three key components: (1) second order natural gradient optimization using K-FAC for stable and efficient policy updates, (2) a margin-aware gradient manipulation mechanism that dynamically balances reward and cost objectives based on safety status, and (3) a minibatch level KL rollback strategy that enforces conservative, trust region style updates. Empirical results on the Safety Gymnasium benchmark demonstrate that this combination effectively maintains constraint satisfaction across diverse tasks and agent types, even under limited model capacity or high observation complexity. The synergy between these components enables consistently safe policy learning, with KFCPO outperforming existing baselines in balancing return and constraint satisfaction, especially in scenarios where other methods either violate constraints or exhibit instability.

\textbf{Future Work.} Although this study primarily focused on demonstrating safety and robustness, several performance related aspects remain unexplored. Due to hardware limitations, we restricted our evaluation to the shallow default two-layer networks, which prevented us from exploring the full potential of K-FAC in deeper architectures where its layer wise parallelism is more beneficial. Likewise, in Button tasks, the relatively high state dimensionality compared to model capacity likely constrained policy expressiveness, thereby hindering reward learning. A more comprehensive evaluation should include deeper and wider networks to better assess KFCPO's scalability and performance under more demanding conditions.

\end{document}